\documentclass[conference]{IEEEtran}
\usepackage{times}

\usepackage{algorithm}
\usepackage[noend]{algpseudocode}
\usepackage[square, numbers, sort&compress]{natbib}
\usepackage[font=small, labelfont=bf]{caption}
\usepackage{multicol}
\usepackage[bookmarks=true]{hyperref}
\usepackage{lipsum}
\usepackage{xspace}
\usepackage{todonotes}
\usepackage{amsmath}
\usepackage{amssymb}
\usepackage{booktabs}

\def\eqref#1{equation~\ref{#1}}

\def\1{\bm{1}}

\DeclareMathAlphabet{\mathsfit}{\encodingdefault}{\sfdefault}{m}{sl}
\SetMathAlphabet{\mathsfit}{bold}{\encodingdefault}{\sfdefault}{bx}{n}

\def\gB{{\mathcal{B}}}

\def\gL{{\mathcal{L}}}

\def\gT{{\mathcal{T}}}

\def\gW{{\mathcal{W}}}

\newcommand{\E}{\mathbb{E}}

\hypersetup{
 colorlinks=true,
 citecolor=green,
 linkcolor=black,
 urlcolor=blue}

\begin{document}

\title{\huge \algoName: Vision-Based Kilometer-Scale Navigation \\ with Geographic Hints}

\newcommand{\algoName}[0]{ViKiNG\xspace}
\newcommand{\algoAstar}{\algoName-A$^*$\xspace}
\newcommand{\Astar}{A$^*$\xspace}%

\author{Dhruv Shah, Sergey Levine \\ UC Berkeley 
}

\makeatletter
\let\@oldmaketitle\@maketitle%
\renewcommand{\@maketitle}{\@oldmaketitle%
    \centering
    \includegraphics[width=\linewidth]{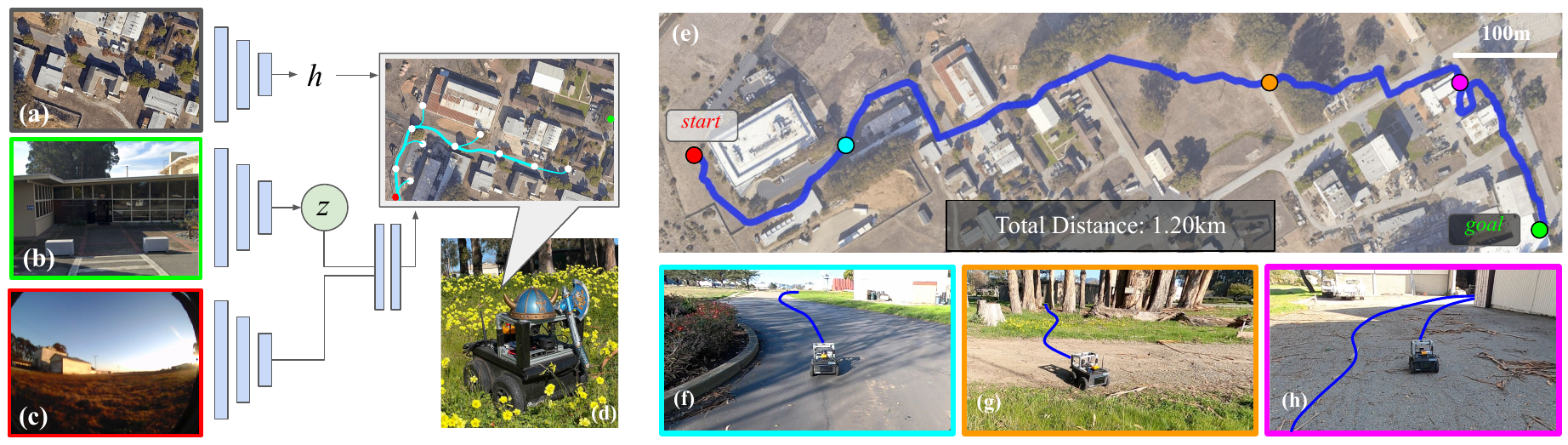}
    \vspace{-0.2in}
    \captionof{figure}{\textbf{Kilometer-scale autonomous navigation with \algoName:} Our learning-based navigation system takes as input the current egocentric image (c), a photograph of the desired destination (b), and an overhead map (which may be a schematic or satellite image) (a) that provides a \emph{hint} about the surrounding layout. The robot (d) uses learned models trained in \emph{other} environments to infer a path to the goal (e), combining local traversability estimates with global heuristics derived from the map. This enables \algoName to navigate \emph{previously unseen} environments (e), where a single traversal might involve following roads (f), off-road driving under a canopy (g), and backtracking from dead ends (h).}
    \label{fig:teaser}
}
\algrenewcommand\algorithmiccomment[2][\normalsize]{{#1\hfill\(\triangleright\) #2}}
\makeatother
\maketitle
\IEEEpeerreviewmaketitle
\setcounter{figure}{1}

\begin{abstract}
Robotic navigation has been approached as a problem of 3D reconstruction and planning, as well as an end-to-end learning problem. However, long-range navigation requires both planning and reasoning about local traversability, as well as being able to utilize general knowledge about global geography, in the form of a roadmap, GPS, or other side information providing important cues. In this work, we propose an approach that integrates learning and planning, and can utilize side information such as schematic roadmaps, satellite maps and GPS coordinates as a planning heuristic, without relying on them being accurate. Our method, ViKiNG, incorporates a local traversability model, which looks at the robot's current camera observation and a potential subgoal to infer how easily that subgoal can be reached, as well as a heuristic model, which looks at overhead maps for hints and attempts to evaluate the appropriateness of these subgoals in order to reach the goal. These models are used by a heuristic planner to identify the best waypoint in order to reach the final destination. Our method performs no explicit geometric reconstruction, utilizing only a topological representation of the environment. Despite having never seen trajectories longer than 80 meters in its training dataset, ViKiNG can leverage its image-based learned controller and goal-directed heuristic to navigate to goals up to 3 kilometers away in previously unseen environments, and exhibit complex behaviors such as probing potential paths and backtracking when they are found to be non-viable. ViKiNG is also robust to unreliable maps and GPS, since the low-level controller ultimately makes decisions based on egocentric image observations, using maps only as planning heuristics. For videos of our experiments, please check out our project page: \href{https://sites.google.com/view/viking-release}{\texttt{sites.google.com/view/viking-release}}
\end{abstract}

\section{Introduction}
\label{sec:intro}
Robotic navigation has conventionally been approached as a geometric problem, where the robot constructs a 3D model of the environment and then plans a path through this model. End-to-end learning-based methods offer an alternative approach, where the robot learns to correlate observations with traversability information directly from experience, without full geometric reconstruction~\citep{zhu2016targetdriven, chen2015deepdriving, kahn2018gcg}. This can be advantageous because, in many cases, geometry alone is neither necessary nor sufficient to traverse an environment, and a learning-based method can acquire patterns that are more directly indicative of traversability, for example by learning that tall grass is traversable~\citep{kahn2020badgr} while seemingly traversable muddy soil should be avoided. More generally, such methods can learn about common patterns in their environment, such as that houses tend to be rectangular, or that fences tend to be straight. These patterns can lead to common-sense inferences about which path should be taken through an unknown environment even before that environment has been fully mapped out~\citep{narasimhan2020scene}.

However, dispensing with geometry entirely may also be undesirable: the spatial organization of the world provides regularities that become important for a robot that needs to traverse large distances to reach its goal. In fact, when humans navigate new environments, they make use of both geographic knowledge, obtained from overhead maps or other cues, and learned patterns~\citep{wiener2009taxonomy}. But in contrast to SLAM, humans don't require maps or auxiliary signals to be very accurate: a person can navigate a neighborhood using a schematic that roughly indicates streets and houses, and reach a house marked on it. Humans do not try to accurately reconstruct geometric maps, but use approximate ``mental maps'' that relate landmarks to each other topologically~\citep{Foo2005humans}. Our goal is to devise learning-enabled methods that similarly make use of geographic \emph{hints}, which could take the form of GPS, roadmaps, or satellite imagery, without requiring these signals to be perfect.

\begin{figure}
    \centering
    \includegraphics[width=0.9\columnwidth]{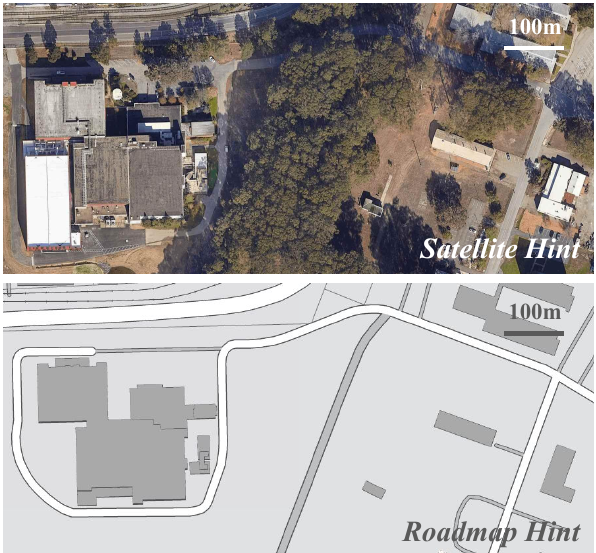}
    \caption{\textbf{Geographic hints} used by \algoName. We evaluate our method with either satellite images or schematic roadmaps, though the approach could be used with any other information of this form, such as contour maps.}
    \label{fig:hints_example}
    \vspace*{-0.2in}
\end{figure}

We consider the problem of navigation from raw images in a \emph{novel} environment, where the robot is tasked with reaching a user-designated goal, specified as an egocentric image, as shown in Figure~\ref{fig:teaser}. Note that the robot has \emph{no prior experience} in the target environment.The robot has access to geographic side information in the form of a schematic roadmap or satellite imagery (see Figure~\ref{fig:hints_example}), which may be outdated, noisy, and unreliable, and approximate GPS. This information, while not sufficient for navigation by itself, contains useful cues that can be used by the robot. The robot also has access to a large and diverse dataset of experience from \emph{other} environments, which it can use to learn general navigational affordances. We posit that an effective way to build such a robotic system is to combine the strengths of machine learning with informed search, by incorporating the geographic hints into a learned heuristic for search.
The robot uses approximate GPS coordinates and an overhead map as geographic side information to help solve the navigation task, but does not assume that this information is particularly accurate---resembling a person using a paper map, the robot uses the GPS localization and an overhead map as \emph{hints} to aid in visual navigation. Note that while we do assume access to GPS, the measurements are only accurate up to 2-5 meters (4-10$\times$ the scale of the robot), and cannot be used for local control.

The primary contribution of this work is 
\algoName, an algorithm that combines elements of end-to-end learning-based control at the low level
with a higher-level heuristic planning method that uses this image-based controller in combination with the geographic hints. The local image-based controller is trained on large amounts of prior data from \emph{other} environments, and reasons about navigational affordances directly from images without any explicit geometric reconstruction. The planner selects candidate waypoints in order to reach a faraway goal,
incorporating the geographic side information as a planning heuristic. Thus, when the hints are accurate, they help the robot navigate toward the goal, and when they are inaccurate, the robot can still rely on its image observations to search through the environment.
We demonstrate \algoName on a mobile ground robot and evaluate its performance in a variety of open-world environments not seen in the training data, including suburban areas, nature parks, and a university campus. Our local controller is trained on 42 hours of navigational data, and we test our complete system in 10 different environments. Despite never seeing trajectories longer than 80 meters in its training data, \algoName can effectively use geographic side information in the form of overhead maps to reach user-specified goals in \emph{previously unseen environments} over 2 kilometers away in under 25 minutes.

\section{Related Work}
\label{sec:related_work}

Robotic navigation has been studied from a number of different perspectives in different fields. Classically, it is often approached as a problem of geometric mapping or reconstruction followed by planning~\citep{Thrun2005}. In unknown environments, the mapping problem can be formulated in terms of information gain with local strategies~\citep{bourgault2002exploration, kollar2008optimization, tabib2019real}, global strategies based on the frontier method~\citep{yamauchi1997frontier, holz2011frontier, charrow2015mapping, Charrow2015planning}, or by sampling ``next-best views''~\citep{Connolly1985nbv, papachristos2017nbv, selin2019nbv}, but such methods typically aim to map or reconstruct an entire environment, rather than achieve a single navigational goal. Active exploration methods have sought to modify this by jointly incentivizing an exploration objective along with reconstruction of the map~\citep{delmerico2017active, lluvia2021active}. Both the goal-directed and mapping-focused methods aim to reconstruct the geometry of their environment, and do not directly benefit from training with prior data.
Some approaches have sought to incorporate learning into mapping and reconstruction~\citep{mccormac2017semantic, murthy2019gradslam}, which benefits from prior data, but still aims at dense geometric reconstruction. Our approach uses a model that is trained with data from prior environments to predict \emph{traversability} rather than geometry, and this model is then used in combination with geographic hints to plan a path to the goal.

In this respect, \algoName is also related to prior work on learning-based navigation, which is often formulated in terms of the ``PointGoal'' task~\citep{savva2019habitat}. Many such works rely on simulation and reinforcement learning, utilizing millions (or billions) of online trials to train a policy~\cite{kumar2018visual, wijmans2020ddppo}. In contrast, our method learns entirely from previously collected offline data, extrapolates to significantly longer paths than it is trained on, and does not require any simulation or online RL. 

A number of prior learning-enabled methods also combine learned models with graph-based planning, using a topological graph to represent the environment~\citep{savinov2018sptm, bruce2018deployable, faust2018prm, shah2020ving, meng2020scaling}. These methods often assume access to data from the test environment to start with a viable graph, which may not be available in a new environment. Some works have studied this unseen setting by predicting explorable areas for semantically rich parts of the environment to accelerate visual exploration~\citep{chaplot2020nts, chaplot2020semantic}. While these methods can yield promising results in a variety of domains, they come at the cost of high sample complexity (over 10M samples)~\citep{savva2019habitat}, making them difficult to use in the real world---the most performant algorithms take 10-20 minutes to find goals up to 50m away~\citep{shah2021rapid}.

The closest prior work to \algoName is by Shah et al. (RECON)~\citep{shah2021rapid}, which uses a learned representation over feasible subgoals to uniformly explore the environment. Like RECON, our method trains a local model that predicts temporal distances and actions for nearby subgoals, and then incorporates this model into a search procedure that incrementally constructs a topological graph in a novel environment. However, in contrast to RECON, which performs an uninformed search, \algoName incorporates geographic hints in the form of approximate GPS coordinates and overhead maps. This enables \algoName to reach faraway goals, up to 25$\times$ further away than the furthest goal reported by RECON, and to reach goals up to 15$\times$ faster than RECON when exploring a novel environment.

\section{Visual Navigation with Geographic Hints}
\label{sec:method}

\begin{figure}
    \centering
    \includegraphics[width=\columnwidth]{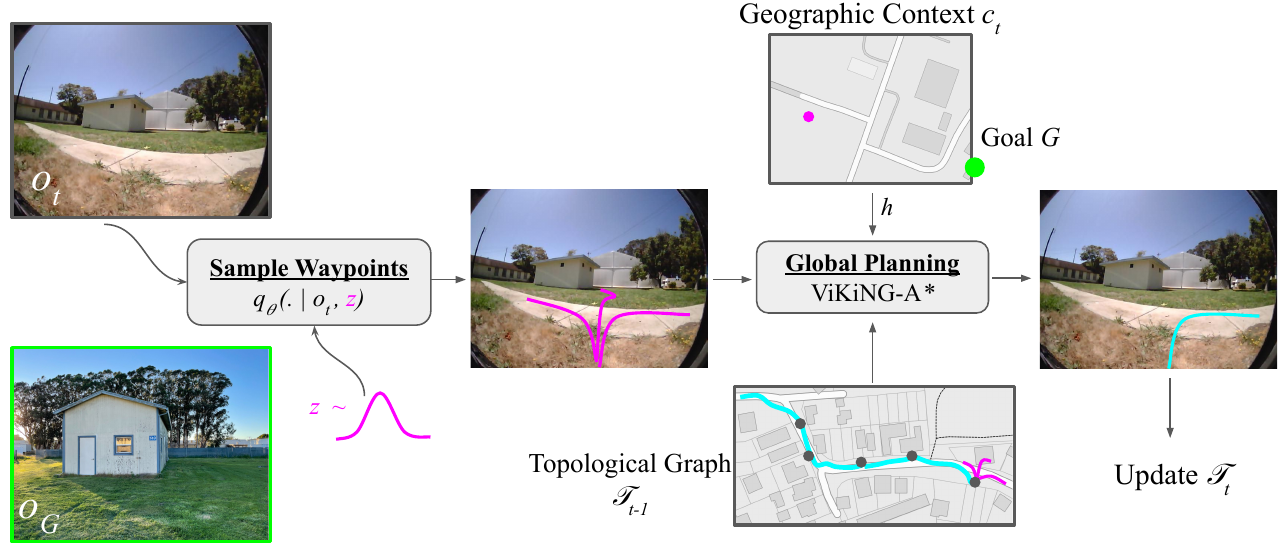}
    \caption{\textbf{An overview of our method.} \algoName uses latent subgoals $z$ proposed by a learned low-level controller, which operates on raw image observations $o_t$, for global planning on a topological graph $\mathcal{T}$ to reach a distant goal $o_G$, indicates by a photograph and an approximate GPS location. A learned heuristic parses the overhead image $c_t$ to bias this search towards the goal.}
    \label{fig:overview}
    \vspace*{-0.2in}
\end{figure}

Our aim is to design a robotic system that learns to use first-person visual observations to reach user-specified landmarks, while also utilizing geographic \emph{hints} in the form of approximate GPS coordinates and overhead maps. At the core of our approach is a deep neural network that takes in the robot's current camera observation $o_t$, as well as an observation $o_w$ of a potential subgoal $w$ (we use ``subgoal'' and ``waypoint'' interchangeably), and predicts the time to reach $w$ (or ``temporal distance''), the best current action to do so, and the resulting spatial offset in terms of GPS coordinates.
This model can also sample latent representations of potential \emph{reachable} waypoints from the current observation $o_t$, which are used as candidate subgoals for planning. The model is trained on large amounts of data from a variety of training environments and, when the robot is placed in a \emph{new} environment that it has not seen, it is used to incrementally construct a \emph{topological} (non-geometric) graph to navigate to a distant user-specified goal. This goal is indicated by a photograph with an approximate GPS coordinate, and may be several kilometers away. The learned model alone is insufficient to navigate to such a distant goal in one shot, and therefore our planner uses a combination of the model's predictions and geographic information to plan a sequence of subgoals that search for a path through the environment, incrementally constructing the graph.

This process corresponds to a kind of heuristic search, where the geographic side information provides a heuristic to bias the robot to explore towards the goal as it constructs the topological graph. The latent goal model is used to determine reachability in this topological graph, and the geographic heuristic is used to steer the graph by exploring the environment. In a novel environment, the robot must incrementally build this graph using \emph{physical search}, by visiting new nodes and expanding its frontier. The decision about \emph{where} to actually go is determined by the first-person images, and the geographic information is used only as a heuristic, allowing \algoName to remain robust to noisy or unreliable side information. We overview our method in Figure~\ref{fig:overview}.

\subsection{Low-level Control with a Latent Goal Model}
\label{sec:representation}

Our low-level model maps the current image observation $o_t$ and a waypoint observation $o_w$ to: (1) the temporal distance $d_t^w$ to reach $w$ from $o_t$; (2) the first action $a_t^w$ that the robot must take now to reach $w$; (3) a prediction of the (approximate) offset in GPS readings between $o_t$ and $w$, $x_t^w$. (1) and (3) will be used by the higher-level planner, and (2) will be used to drive to $w$, if needed. We would also like this model to be able to \emph{propose}, in a learned latent space, potential subgoals $w$ that are reachable from $o_t$, and predict their corresponding values of $d_t^w$, $a_t^w$, and $x_t^w$.

\begin{figure}
    \centering
    \includegraphics[width=\columnwidth]{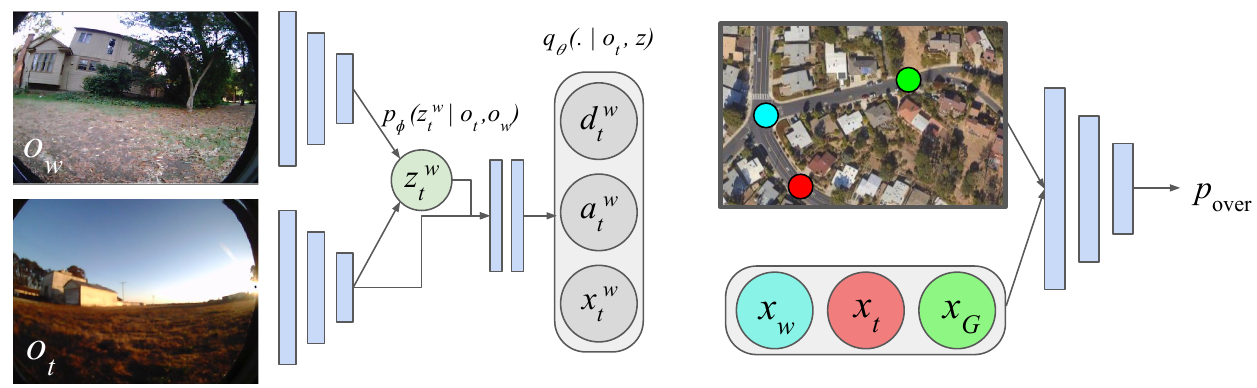}
    \caption{\textbf{The learned models used by \algoName.} The latent goal model (left) takes in the current image $o_t$. It also takes in either a true waypoint image $o_w$, or samples a \emph{latent} waypoint $z_t^w \sim r(z_t^w)$ from a prior distribution, and then predicts, its temporal distance from $o_t$ ($d_t^w$), the action to reach it ($a_t^w$), and its approximate GPS offset ($x_t^w$). The heuristic model (right) takes in an overhead image $c_t$, the approximate GPS coordinates of the current location ($x_t$) and destination ($x_G$), and the coordinates of the waypoint inferred by the latent goal model ($x_w$), and predicts an approximate heuristic value of the waypoint $w$ for reaching the final destination.}
    \label{fig:architecture}
    \vspace*{-0.2in}
\end{figure}

We present the model in Figure~\ref{fig:architecture}, with precise architecture details in the supplementary materials. The model is trained by sampling pairs of time steps in the trajectories in the training set. For each pair, the earlier time step image becomes $o_t$, and the later image becomes $o_w$. The number of time steps between them provides the supervision for $d_t^w$, the action taken at the earlier time step supervises $a_t^w$, and the later GPS reading is transformed into the coordinate frame of the earlier time step to provide supervision for $x_t^w$. The model is trained via maximum likelihood. Note that by training the model on data in this way, we not only enable it to evaluate reachability of prospective waypoints, but also make it possible to inherit behaviors observed in the data. For example, in our experiments, we will show that the model has a tendency to follow sidewalks and forest trails, a behavior it inherits from the portion of the dataset that is collected via teleoperation.

Besides predicting $d_t^w$, $a_t^w$, and $x_t^w$, our planner requires this model to be able to \emph{sample} potential reachabale waypoints from $o_t$ (see Figure~\ref{fig:overview}). We implement this via a variational information bottleneck (VIB) inside of the model that bottlenecks information from $o_w$. Thus, the model can either take as input a real image $o_w$ of a prospective waypoint, or it can sample a \emph{latent} waypoint $z_t^w \sim r(z_t^w)$ from a prior distribution. We train the model so that sampled latent waypoints correspond to feasible locations that the robot can reach from $o_t$ without collision.

\noindent \textbf{Training the latent goal model:} The full model, illustrated in Figure~\ref{fig:architecture}, can be split into three parts: a waypoint encoder $p_\phi(z_t^w | o_w, o_t)$, a waypoint prior $r(z_t^w)$, and a predictor $q_\theta(\{a,d,x\}_t^w | z_t^w, o_t)$. The latent waypoint representation $z_t^w$ can either be sampled from the prior (which is fixed to $r(z_t^w)\triangleq \mathcal N(0, I)$), or from the encoder $p_\phi(z_t^w | o_w, o_t)$ if a waypoint image $o_w$ is provided. This latent waypoint is used together with $o_t$ to predict all desired quantities according to $q_\theta(\{a,d,x\}_t^w | z_t^w, o_t)$. The training set consists of tuples $(o_t, o_w, \{a,d,x\}_t^w)$, but the model must be trained so that samples $z_t^w \sim r(z_t^w)$ also produce valid predictions. We accomplish this by means of the VIB~\citep{alemi2016deep}, which regularizes the encoder $p_\phi(z_t^w | o_w, o_t)$ to produce distributions that are close to the prior $r(z_t^w)$ in terms of KL-divergence. We refer the reader to prior work for a derivation of the VIB~\citep{alemi2016deep}, and present our training objective for $p_\phi$ and $q_\theta$ below:
\begin{align}
\gL_\text{VIB}(\theta,\phi) = E_{\mathcal{D}}[& -\mathbb E_{p_\phi}  \left[ \log q_\theta\big(\{a, d, x\}_t^w \mid z_t^w, o_t\big)\right] \nonumber \\ 
        & +\beta \mathrm{KL}\left(p_\phi(z_t^w \mid o_w, o_t) || r(z_t^w) \right) ] \label{eq:bottleneck_tractable}
\end{align}
The outer expectation over all tuples $(o_t, o_w, \{a,d,x\}_t^w) \in \mathcal{D}$ in the training distribution is estimating using the training set. The first term causes the model to accurately predict the desired information, while the second term forces the encoder to remain consistent with the prior, which makes the model suitable for sampling latent waypoints according to $z_t^w \sim r(z_t^w)$. As the encoder $p_\phi$ and decoder $q_\theta$ are conditioned on $o_t$, the representation $z_t^w$ only encodes \emph{relative} information about the subgoal from the context---this allows the model to represent \emph{feasible} subgoals in new environments, and provides a compact representation that abstracts away irrelevant information, such as time of day or visual appearance. An analogous representation has been proposed in prior work~\citep{shah2021rapid}, but did not predict spatial offsets and was used only for \emph{uninformed} exploration without geographic hints.

\subsection{Informed Search on a Topological Graph}
\label{sec:search_primer}

The model described above can effectively reach nearby subgoals, for example those on which the robot has line of sight, but we wish to reach goals that are more than a kilometer away. To reach distant goals, we combine the model with a search procedure that incorporates geographic hints from satellite images or roadmaps. The system does not require this information to be accurate, instead using it as a planning heuristic while still relying on egocentric camera images for control.
Our high-level planner plans over a topological graph $\gT$ that it constructs incrementally using the low-level model in Section~\ref{sec:representation} as a local planner. We first describe a generic version of the algorithm for any heuristic, and then describe the data-driven heuristic function that we extract from the geographic hints via contrastive learning.

\noindent \textbf{Challenges with \emph{physical search}:}
Our ``search'' process involves the robot physically searching through the environment, and is not purely a computational process. In contrast to standard search algorithms (e.g., Dijkstra, A$^*$, IDA$^*$, D$^*$, etc.), each ``step'' of our search involves the robot driving to a subgoal and updating the graph. Standard graph search algorithms assume (i) the ability to \emph{visit} any arbitrary node, and (ii) access to a set of \emph{neighbors} for every node and the corresponding ``edge weight,'' before visiting each neighbor. Physical search with a robot violates these assumptions, since robots cannot ``teleport'' and \emph{visiting} a node incurs a driving cost. Furthermore, the real world does not provide ``edge weights'' and the robot needs to estimate the cost to reach an unvisited node \emph{before} actually driving to it.

\begin{algorithm}[t]
\caption{\emph{\algoAstar for Physical Search}}
\label{alg:exploration_algorithm}
\begin{algorithmic}[1]
\Function{\algoAstar}{start $S$, goal info $o_G$, $x_G$}
\State $\Omega \leftarrow \{S\}$
\While{$\Omega$ not empty}
    \State $w_t$ $\gets$ min($\Omega$, $f$)
    \label{algline:min}
    \State $\mathrm{DriveTo}(w_t)$
    \Comment{{\footnotesize update visitations $v$ on the way}}
    \label{algline:driveto}
    \State observe image $o_t$
    \label{algline:camera}
    \State add $w_t$ to graph $\mathcal{T}$
    \Comment{{\footnotesize use $q_{\theta,\phi}$ on $o_t$ to get distances}}
    \label{algline:update_graph}
    \State if $\mathrm{close}(o_t,o_G)$ \textbf{finish}
    \Comment{{\footnotesize use $q_{\theta,\phi}(\{a, d, x\}_t^w | o_t, o_G)$}}
    \label{algline:success}
    \State remove $w_t$ from $\Omega$
    \State sample waypoints $w$ near $w_t$ (Section~\ref{sec:representation})
    \label{algline:sample_latentgoals}
    \For{each $w$ sampled near $w_t$}
        \State \textbf{if} not $\mathrm{contains}(\Omega,w)$ \textbf{then} add $w$
        \label{algline:add_to_openset}
    \EndFor
    \For{each waypoint $w \in \Omega$}
    \label{algline:update_all_openset}
        \State $f(w) = g(t,w) + d^w_{\text{Pr}[w]} + h(w) + v(\text{Pr}[w])$
        \label{algline:update_f}
    \EndFor
\EndWhile
\State \textbf{return} failure

\EndFunction
\end{algorithmic}
\end{algorithm}
\noindent \textbf{An algorithm for informed physical search:}
To solve these challenges, we design \algoAstar, an  A$^*$-like search algorithm that uses our latent goal model and a learned heuristic to perform \emph{physical search} in real-world environments. While \algoAstar does prefer shorter paths, it does not aim to be optimal (in contrast to A$^*$), only to reach the goal successfully. We will use a heuristic $h(w)$, fully described in the next section, which we assume provides a \emph{comparative} evaluation of candidate waypoints in terms of their anticipated temporal distance to the destination. Algorithm~\ref{alg:exploration_algorithm} outlines \algoAstar.

Like \Astar, \algoAstar maintains a priority queue ``open set'' $\Omega$ of unexplored fringe nodes and a ``current'' node that represents the least-cost node in this set, which we refer to as $w_t$. It also maintains a graph with visited waypoints, $\mathcal{T}$, where nodes correspond to images seen at those nodes, and edges correspond to temporal distances estimated by the model in Section~\ref{sec:representation}. At every iteration, the robot \emph{drives} to the least-cost node in the open set (L\ref{algline:driveto}), using a procedure that we outline later. When it reaches $w_t$, it observes the image $o_t$ using its camera (L\ref{algline:camera}). This allows it to add $o_t$ to the graph $\mathcal{T}$ (L\ref{algline:update_graph}), connecting it to other nodes by evaluating the distances using the model in Section~\ref{sec:representation}. The graph construction is analogous to prior work~\citep{shah2020ving,shah2021rapid}. If the robot is close to the final goal image $o_G$ according to the model (L\ref{algline:success}), the search ends. $o_t$ also allows it to sample nearby candidate waypoints using the model in Section~\ref{sec:representation} (L\ref{algline:sample_latentgoals}): first sampling $z_t^w \sim r(z_t^w)$ from the prior, and then decoding distances $d_t^w$, $a_t^w$, and $x_t^w$, from which it can compute absolute locations as $x_w = x_t + x_t^w$. Each sampled waypoint is stored in the open set, and annotated with the current image $o_t$ and $d_t^w$. We refer to $w_t$ as the \emph{parent} of $w$, and index it as $\text{Pr}[w]$. Note that we do \emph{not} have access to the image $o_w$, as we have not visited the sampled waypoint $w$ yet, and therefore we must store the current image $o_t$ instead. This also means that we \emph{cannot} connect these waypoints to the graph $\mathcal{T}$ except through their parent. Next, we re-estimate the cost of each waypoint in the open set, including the newly added waypoints.

The cost for each waypoint $w \in \Omega$ from the current point $w_t$ consists of four terms (L\ref{algline:update_f}): (1) $g(t,w)$, the cost to navigate to the \emph{parent} of $w$, which is part of the graph $\mathcal{T}$; this can be computed as a shortest path on the graph $\mathcal{T}$, and is zero for the current node. (2) $d_{\text{Pr}[w]}^w$, the distance from the parent of $w$ to $w$ itself. (3) $h(w)$, the heuristic cost estimate of reaching the final goal from $w$ (see Section~\ref{sec:heuristic}). (4) $v(\text{Pr}[w])$, the visitation count of $\text{Pr}[w]$, computed as $C N(\text{Pr}[w])$, where $C$ is a constant and $N(\text{Pr}[w])$ is a count of how many times the robot drove to $\text{Pr}[w]$ via the DriveTo subroutine; this acts as a \emph{novelty bonus} to encourage the robot to explore novel states, a strategy widely used in RL~\citep{lai1985ucb, bellemare2016unifying}. Summing these terms expresses a preferences for nodes that are fast to reach from $w_t$ (1 + 2), get us closer to the goal (3), and have not been heavily explored before (4). At the next iteration (L\ref{algline:min}), the robot picks the lowest-cost waypoint and again drives to it.

To navigate to a selected waypoint $w$ (DriveTo), the robot employs a procedure analogous to prior work on learning-based navigation with topological graphs~\citep{shah2020ving,shah2021rapid}, planning the shortest path through $\mathcal{T}$, and selecting the next waypoint on this path. Once the waypoint $w$ is selected, the model $q_{\theta,\phi}(\{a, d, x\}_t^w | o_t, o_w)$ is used to repeatedly choose the action $a_t^w$ based on the current image $o_t$, until the distance $d_t^w$ becomes small, indicating that the waypoint is reached and the robot can navigate to the next waypoint (in practice, it's convenient to replan the path at this point, as is standard in MPC). Each time the DriveTo subroutine reaches a node, it also increments its count $N(w)$ which is used for the novelty bonus $v(w)$. The helper function $\mathrm{close}$ uses the model in Section~\ref{sec:representation} to check if the estimated temporal distance $d_t^w$ is less than $\epsilon$ for two observations, and the $\mathrm{contains}$ operation on a set checks if a given node is \emph{close} to any node inside the set. These modifications allow \Astar-like operations on the nodes of our graph, which are continuous variables.

\subsection{Learning a Goal-Directed Heuristic for Search}
\label{sec:heuristic}

We now describe how we extract a heuristic $h$
from geographic side information. As a warmup, first consider the case where we only have the GPS coordinates for a waypoint ($x_w$) and final goal ($x_G$). We can use $\|x_g - x_w\|$ as a heuristic to bias the search to waypoints in the direction of the goal, and this heuristic can be readily obtained from the model in Section~\ref{sec:representation}.
However, we would like to compute the heuristic function using some  side information $c_t$, such as a roadmap or satellite image, that does not lie in a metric space. Thus, we need to \emph{learn} the heuristic function from data. Since \algoAstar does not aim to be optimal (only seeking a feasible path), we do not require the heuristic to be admissible.

We train the heuristic $h_\text{over}(x_w, x_G, x_t, c_t)$
to score the favorability of a sampled candidate waypoint $w$ for reaching the goal $G$ from current location $x_t$, given side information $c_t$. In our case, $c_t$ is an overhead image that is roughly centered at the current location of the robot.
Our heuristic is based on an estimator for the probability $p_\text{over} (w \rightarrow G | x_w, x_G, x_t, c_t)$ that a given waypoint $w$ lies on a valid path to the goal $G$.
We use the same training set as in Section~\ref{sec:representation} to learn a predictor for $p_\text{over}$. Given $p_\text{over}$, we can generate a heuristic $h_\text{over} := \lambda_\text{over}(1 - p_\text{over})$ to steer \algoAstar towards the goal (Alg.~\ref{alg:exploration_algorithm} L\ref{algline:update_all_openset}). Note that, since we evaluate the heuristic for sampled candidate waypoints, we do not have access to $x_w$, but we can predict it by using the model in Section~\ref{sec:representation} to infer the offset $x_t^w$ using $o_t$ and the sampled latent code, and then calculate $x_w$ from $x_t$ and $x_t^w$. Thus, the heuristic is technically a function of $c_t$, $o_t$, $x_t$, and $x_G$.

Our procedure for training $p_\text{over} (w \rightarrow G | x_w, x_G, x_t, c_t)$ is based on InfoNCE~\citep{oord2019representation}, a contrastive learning objective that can be seen as a binary classification problem between a set of \emph{positives} and \emph{negatives}. At each training iteration, we sample a random batch $\gB$ of sub-trajectories $k$ from our training set, where $x_S$ is the start of $k$ and $x_E$ is the end, and $c_S$ is an overhead image centered at $x_S$. We sample a positive example by picking a random time step in this subtrajectory, and using its position $x_{w^+}$. The negatives $x_{w^-}$ are locations of other randomly sampled time steps from other trajectories, comprising the set $\gW^-$.
In this way, we train a neural network model to represent $p_\text{over} (w \rightarrow G | x_w, x_G, x_t, c_t)$ (see Figure~\ref{fig:architecture}, right) via the InfoNCE objective:
\begin{equation}
    \gL_\text{NCE} = - \E_\gB \left[\log \frac {p_\text{over}(w^+ \rightarrow E | x_{w^+}, x_E, x_S , c_S)} {\sum_{w^- \in \gW^-} p_\text{over}(w^- \rightarrow E | x_{w^-}, x_E, x_S , c_S)}\right]
    \label{eq:nce_objective}
\end{equation}
This heuristic can only reason about waypoints and goals at the scale of individual trajectories in the training set (up to 50m). For kilometer-scale navigation, the heuristic needs to make predictions for goals that are much further away, so we take inspiration from goal chaining in reinforcement learning~\citep{chebotar2021actionable} and combine overlapping trajectories in the training set (according to GPS positions) into larger trajectory groups. For a batch $\gB$ of trajectories, we combine two trajectories if they intersect in 2D space. The resulting macro-trajectories thus have multiple start and goal positions, and can extend for several kilometers. We then sample the sub-trajectories for $x_S$, $x_E$, and $x_{w^+}$ from these much longer macro-trajectories, giving us positive examples between very distant $x_S$, $x_E$ pairs. This allows $p_\text{over}$ to be trained on a vast pool of long-horizon goals and improves the reliability of the heuristic. We provide more details about this procedure in Appendix~\ref{app:implementation_details}.

\section{\algoName in the Real World}
\label{sec:evaluation}

\begin{figure*}
    \centering
    \includegraphics[width=0.95\linewidth]{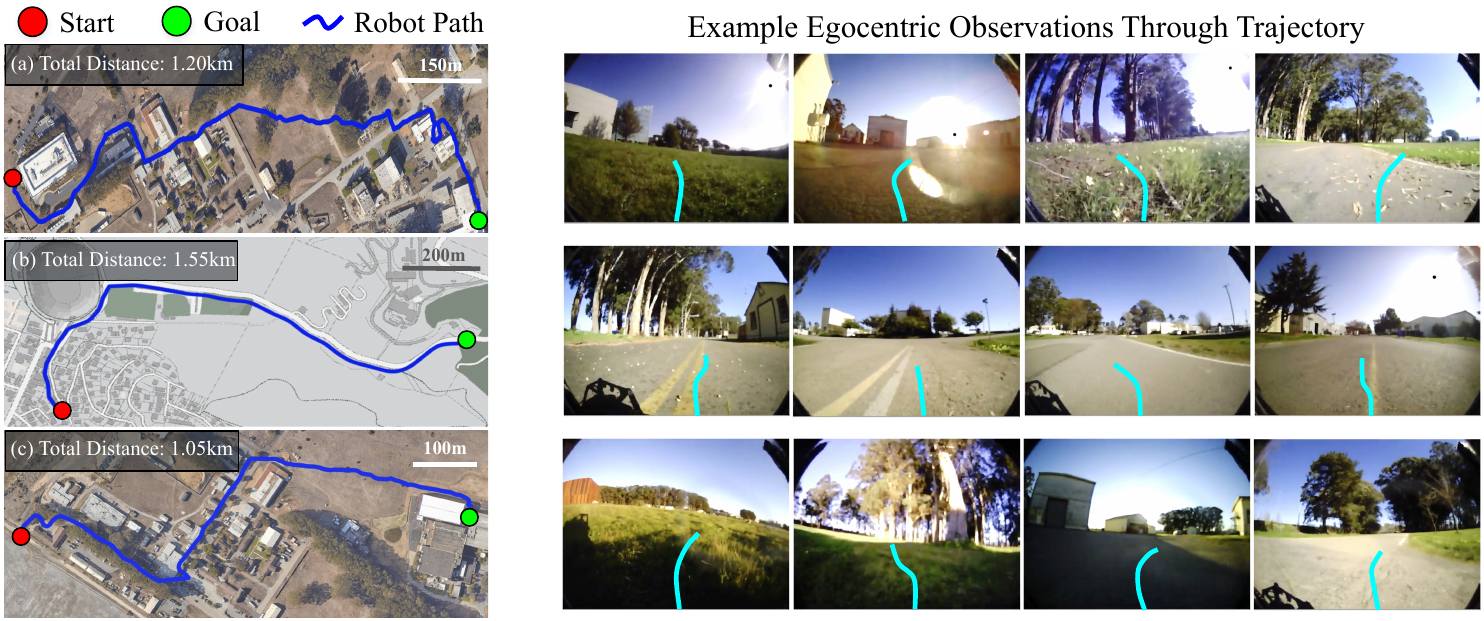}
    \caption{Examples of kilometer-scale goal-seeking in \emph{previously unseen} environments using only egocentric images (right) and a schematic roadmap or satellite image as \emph{hints} (left). \algoName can navigate in complex environments composed of roads, meadows, trees and buildings.}
    \label{fig:qualitative_longrange}
    \vspace*{-0.1in}
\end{figure*}

We now describe our experiments deploying \algoName in a variety of real-world outdoor environments for kilometer-scale navigation. Our experiments compare \algoName to other learning-based methods, evaluate its performance at different ranges, and study how it responds to degraded or erroneous geographic information.

\subsection{Mobile Robot Platform}
We implement \algoName on a Clearpath Jackal UGV platform (see Fig.~\ref{fig:teaser}). The default sensor suite consists of a 6-DoF IMU, a GPS unit for approximate global position estimates, and wheel encoders to estimate local odometry. Under open skies, the GPS unit is accurate up to 2-5 meters, which is 4-10$\times$ the size of the robot. In addition, we added a forward-facing $170^\circ$ field-of-view RGB camera. Compute is provided by an NVIDIA Jetson TX2 computer, and a cellular hotspot connection provides for monitoring and (if necessary) teleoperation. Our method uses only the monocular RGB images from the onboard camera, unfiltered measurements from onboard GPS, and overhead images (roadmap or satellite) queried at the current GPS location, without any other processing.

\subsection{Offline Training Dataset}
\label{sec:offline_dataset}

Our aim is to leverage data collected in a wide range of different environments to (i) enable the robot to learn navigational affordances that generalize to novel environments, and (ii) learn a global planning heuristic to steer physical search in novel environments. To create a diverse dataset capturing a wide range of navigation behavior, we use 30 hours of publicly available robot navigation data collected using an autonomous, randomized data collection procedure in office park style environments~\citep{shah2021rapid}. We augmented this dataset with another 12 hours of teleoperated data collected by driving on city sidewalks, hiking trails, and parks. 
Notably, \algoName never sees trajectories longer than 80 meters, but is able to leverage the learned heuristic (Section~\ref{sec:heuristic}) to reach goals over a kilometer away at over 80\% of the average speed in the training set. The average trajectory length in the dataset is 45m, whereas our experiments evaluate runs in excess of 1km. The average velocity in the dataset is 1.68 m/s, and the average velocity the robot maintains in testing is 1.36 m/s. We provide more details about the dataset in Appendix~\ref{app:dataset}.

\subsection{Kilometer-Scale Testing}
\label{sec:deployment}

For evaluation, we deploy \algoName in a variety of \emph{previously unseen} open-world environments to demonstrate kilometer-scale navigation. Figure~\ref{fig:qualitative_longrange} shows the path taken by the robot in search for a user-specified goal image and location.
\algoName is able to utilize geographic hints, in the form of a roadmap or satellite image centered at its current position, to steer its search of the goal. In a university campus (Fig.~\ref{fig:qualitative_longrange}(a, c)), we observe that the robot can identify large buildings along the way and plan around it, rather than following a greedy strategy.
Since the training data often contains examples of the robot driving around buildings, \algoName is able to leverage this prior experience and generalize to novel buildings and environments. On city roads (Fig.~\ref{fig:qualitative_longrange}(b)), the learned heuristic shows preference towards following the sidewalks, a characteristic of the training data in city environments. It is important to note that while the robot has seen some prior data on sidewalks and in suburban neighborhoods, it has never seen the specific areas (see Appendix~\ref{app:dataset} for further details). For videos of our experiments, please check out our project page. 

These long-range experiments also exhibit successful backtracking behavior---when guided into a cul-de-sac by the planner,
\algoName turns around and resumes its search for the goal from another node in the ``openSet'', reaching the goal successfully (see Figure~\ref{fig:teaser}(h)). While the learned heuristic provides high-level guidance, the local control is done solely from first person images. This is illustrated in Figure~\ref{fig:teaser}(g), where the robot navigates through a forest, where the satellite image does not contain any useful information about navigating under a dense canopy. \algoName is able to successfully navigate through a patch of trees using the image-based model described in Section~\ref{sec:representation}. We can also provide \algoName with a set of goals to execute in a sequence to provide more guidance about the path (e.g., an inspection task with landmarks), as demonstrated in the next experiment.

\begin{figure}
    \centering
    \includegraphics[width=0.9\columnwidth]{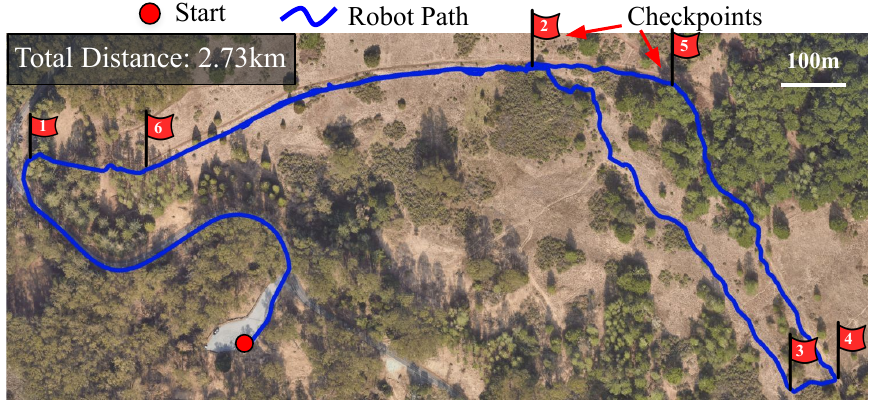}
    \caption{\algoName can follow a sequence of goal checkpoints to perform search in complex environments, such as this 2.73km hiking trail.}
    \label{fig:qualitative_hiking}
    \vspace*{-0.1in}
\end{figure}

\begin{figure}
    \centering
    \includegraphics[width=0.9\columnwidth]{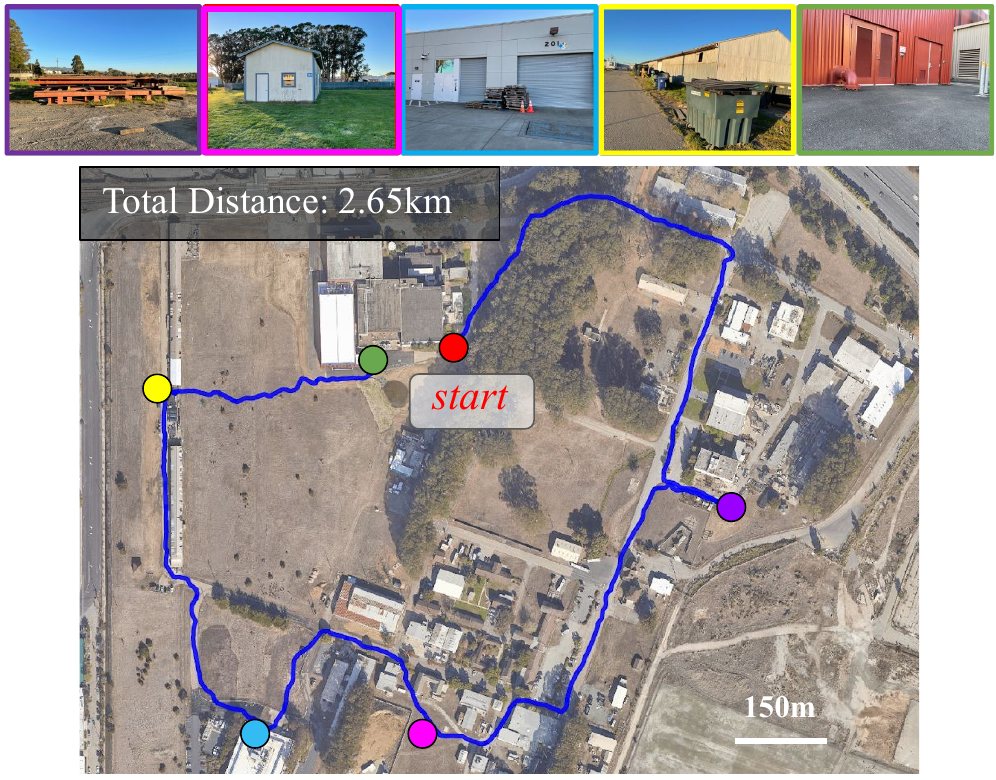}
    \caption{\algoName can utilize a satellite image to follow a sequence of visual landmarks (top) in complex suburban environments, such as this 2.65km loop stretching across buildings, meadows and roads.}
    \label{fig:inspection}
    \vspace*{-0.1in}
\end{figure}

\noindent \textbf{A hiking \algoName:} We deploy \algoName, with access to satellite images as hints, on a 2.7km hiking trail with a 70m elevation gain by providing a sequence of six checkpoint images and their corresponding GPS coordinates. Algorithmically, we run \algoAstar on every goal (one at a time) while reusing the topological graph $\gT$ across goals. Figure~\ref{fig:qualitative_hiking} shows a top-down view
of the path taken by the robot---\algoName is able to successfully combine the strengths of a learned controller for collision-free navigation with a learned heuristic that utilizes the satellite images to encourage on-trail navigation between checkpoints. Since the offline dataset contains examples of trail-following, the robot learns to stay on trails when possible. This behavior is emergent from the data---there is no other mechanism that encourages staying on the trails, and in several cases, a straight-line path between the goal waypoints would not stay on the trail (e.g., the first checkpoint in Figure~\ref{fig:qualitative_hiking}).

\noindent \textbf{Autonomous visual inspection:} We further deploy \algoName in a suburban environment for the task of visual inspection specified by five images of interest. \algoName is able to successfully navigate to the landmarks by using satellite imagery, traveling a distance of 2.65km without any interventions. Figure~\ref{fig:inspection} shows the specified images and a top-down view of the path taken by the robot on the trail.

\subsection{Quantitative Evaluation and Comparisons}
\label{sec:baselines}

We compare \algoName to four prior approaches, each trained using the same offline data as our method. All methods have access to the egocentric images, GPS location, and satellite images, and control the robot via the same action space, corresponding to linear and angular velocities. 

    \noindent \textbf{Behavioral Cloning:} A goal-conditioned behavioral cloning (BC) policy trained on the offline dataset that maps the three inputs to control actions~\citep{Codevilla2017, shah2020ving}.
    
    \noindent \textbf{PPO:} A policy gradient algorithm that maps the three inputs to control actions. This comparison is representative of state-of-the-art ``PointGoal'' navigation in simulation~\citep{wijmans2020ddppo}.
    
    \noindent \textbf{GCG:} A model-based algorithm that uses a predictive model to plan a sequence of actions that reach the goal without causing collision~\citep{kahn2018gcg}. We use GCG in the goal-directed mode with a GPS target, using the onboard camera and satellite images as input modalities.
    
    \noindent \textbf{RECON-H:} A variant of RECON, which uses a latent goal model to represent reachable goals and plans over sampled subgoals to explore a novel environment~\citep{shah2021rapid}. We modify the algorithm to additionally accept the GPS and satellite images as additional inputs alongside the onboard camera image.

\begin{figure}
    \centering
    \includegraphics[width=0.9\columnwidth]{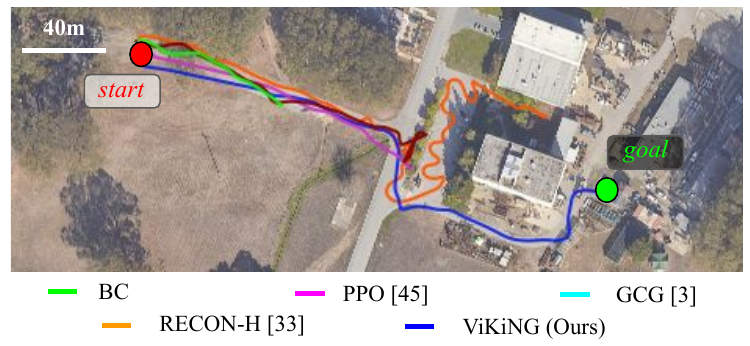}
    \caption{Trajectories taken by the methods in a \emph{previously unseen} environment. Only \algoName is able to effectively use the overhead images to reach the goal (270m away) successfully, following a smooth path around the building. RECON-H and GCG get stuck, while PPO and BC result in collisions.}
    \label{fig:qualitative_baselines}
\end{figure}

We evaluate the ability of \algoName to discover visually-indicated goals in 10 \emph{unseen} environments of varying complexity. For each trial, we provide an RGB image of the desired target and its rough GPS location (accurate up to 5 meters). A trial is marked successful if the robot reaches the goal without requiring a human disengagement (due to a collision or getting stuck). We report the success rates of all methods in these environments in Table~\ref{tab:quantitative_success} and visualize overhead plots of the trajectories in one such environment in Figure~\ref{fig:qualitative_baselines}.

\algoName outperforms all the prior methods, successfully navigating to goals that are over up to 500 meters away in our comparisons, including instances where no other method succeeds. RECON-H is the most performant of the other methods, successfully reaching most goals in the easier environments. Visualizing the robot trajectories (Fig.~\ref{fig:qualitative_baselines}) reveals that RECON-H is unable to successfully utilize the geographic hints and explores greedily on encountering an obstacle. It also gets stuck and is unable to backtrack in 2/10 instances. While GCG also performs well in simpler environments, it is limited by its planning horizon (up to 5 seconds) and gets stuck. PPO and BC both are both unable to learn from prior data and produce collisions with bushes and a parked car, respectively. In contrast, \algoName is able to effectively use the local controller to avoid the obstacles and reach the goal.

Analyzing the performance in the harder
tasks with ranges of up to 500 meters (Table~\ref{tab:quantitative_disengagements}), the average displacements and velocities before a user disengagement (due to collision or getting stuck)
during these runs further confirm that \algoName is able to effectively use the geographic hints to steer the search without running into obstacles. While RECON-H manages to reach some faraway goals, it takes a greedy
path to do so and is over 3$\times$ slower than \algoName (see Fig.~\ref{fig:qualitative_baselines}).

\begin{table}
\centering
{\footnotesize
\begin{tabular}{lccc}
\toprule
Method & Easy & Medium & Hard \\
 & $<50$m & $50-150$m & {$150-500$m} \\ \midrule
Behavior Cloning & 2/3 & 1/4 & 0/3 \\
Offline PPO~\citep{schulman2017ppo} & 2/3 & 1/4 & 0/3 \\
GCG~\citep{kahn2018gcg} & \textbf{3/3} & 2/4 & 0/3 \\
RECON-H~\citep{shah2021rapid} & \textbf{3/3} & 3/4 & 1/3 \\
\textbf{\algoName (Ours)}   & \textbf{3/3} & \textbf{4/4} & \textbf{3/3} \\
\bottomrule
\end{tabular}}
\caption{Comparison of goal-seeking performance against baselines. \algoName successfully reaches all goals. RECON-H and GCG succeed in simpler cases but are unable to utilize the hints effectively for distant goals. PPO and BC fail in all but the simplest cases.}
\label{tab:quantitative_success}
\end{table}

\begin{table}
\centering
{\footnotesize
\begin{tabular}{lcc}
\toprule
Method & Avg. Displacement (m) & Avg. Velocity (m/s) \\ \midrule
Behavior Cloning & 19.5 & 0.35 \\
Offline PPO~\citep{schulman2017ppo} & 47.2 & 0.85 \\
GCG~\citep{kahn2018gcg} & 78.3 & \textbf{1.40} \\
RECON-H~\citep{shah2021rapid} & 188.3  & 0.41 \\
\textbf{\algoName (Ours)}   & \textbf{250.0+} & \textbf{1.36} \\
\bottomrule
\end{tabular}}
\caption{Average robot displacement and velocity before disengagement. \algoName successfully reaches all goals without requiring any disengagements. RECON-H also reaches some distant goals, but the low avg. velocity suggests that it takes an efficient path.}
\label{tab:quantitative_disengagements}
\vspace*{-0.2in}
\end{table}

\section{The Role of Geographic Hints}
\label{sec:imperfect_hints_section}

In this section, we closely examine the role of geographic hints on the performance of \algoName by studying how it deals with a low-fidelity roadmap (versus a satellite image), and with incorrect hints and degraded geographic information. For the experiment in Section~\ref{sec:compare_roadmap}, we use models trained on the same dataset, but using schematic roadmaps as geographic hints. In Sections \ref{sec:outdated_hints} and \ref{sec:invalid_hints}, we use the same satellite image model from Section IV, with no additional retraining to accommodate missing or imperfect geographic information.

\subsection{Comparing Different Types of Hints}
\label{sec:compare_roadmap}

\begin{figure}
    \centering
    \includegraphics[width=\columnwidth]{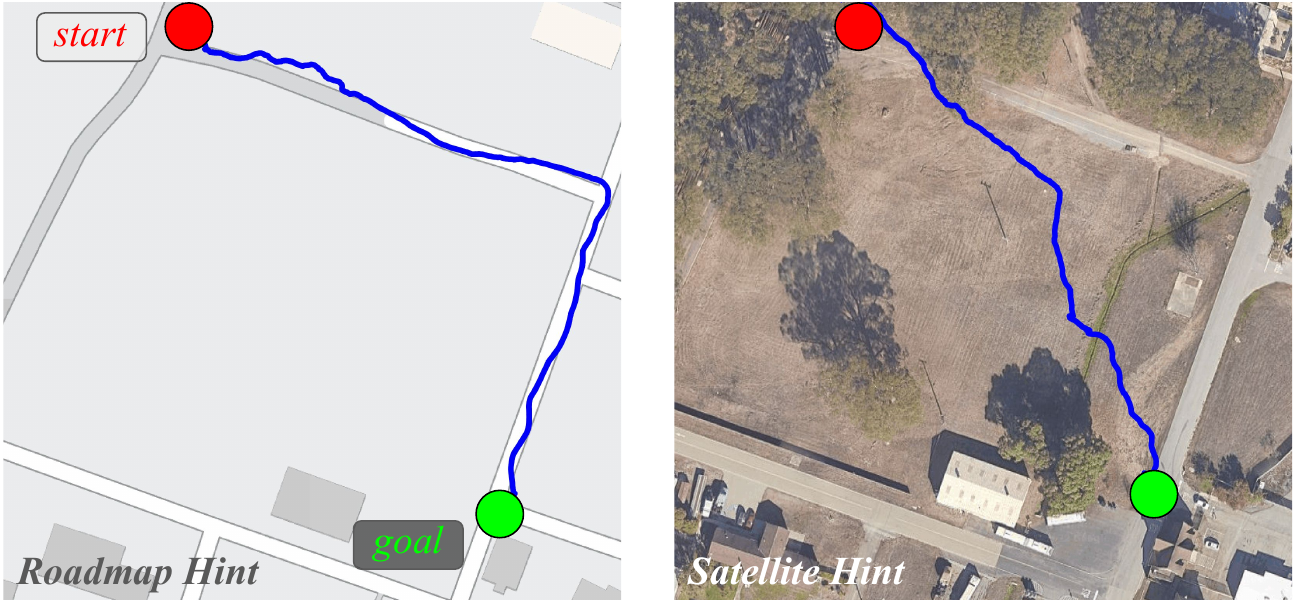}
    \caption{\algoName can use geographic hints in the form of a schematic roadmap or a satellite image. Providing roadmap hints encourages \algoName to follow marked roads (left); with satellite images, it is able to find a more direct path by cutting across a meadow (right).}
    \label{fig:compare_roadmap}
\end{figure}

To understand the nature of hints learned by the heuristic for different sources of geographic side information, we compare two separate versions of \algoName: one trained with schematic roadmaps as hints, and another trained with satellite images. Note that the method is identical in both cases, only the hint image in the data changes. For identical start-goal pairs, we observe that a model trained with roadmaps prefers following marked roads, whereas one trained with satellite images often cuts across patches of traversable terrain (e.g., grass meadows or trails) to take the quicker path, despite being trained on the same data. We hypothesize that this is due to the ability of the learned models to extract better correlations from the feature-rich satellite images, in contrast to the more abstract roadmap. Figure~\ref{fig:compare_roadmap} shows a top-down view of the paths taken by the robot in the two cases in one such experiment.

\subsection{Outdated Hints}
\label{sec:outdated_hints}
\begin{figure}
    \centering
    \includegraphics[width=\columnwidth]{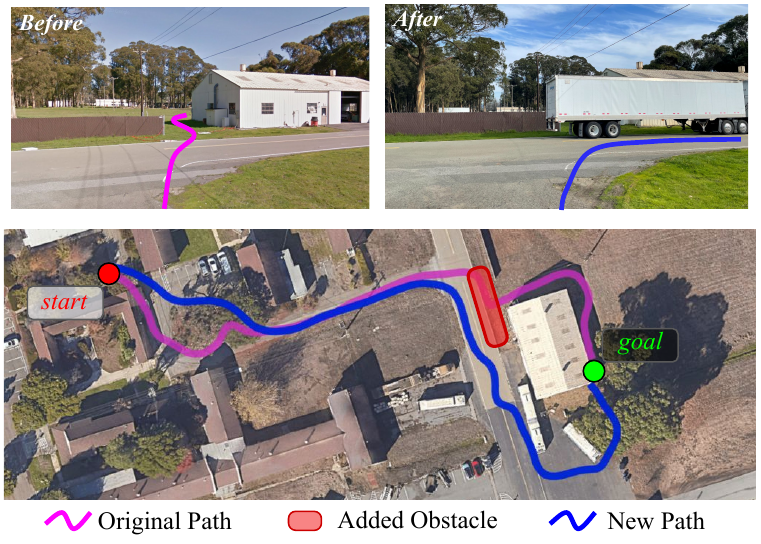}
    \caption{On navigating with outdated hints, like the truck (top right) that is absent in the satellite image, \algoName uses its learned local controller to propose feasible subgoals that avoid obstacles and finds a new path (blue) to the goal that avoids the truck.}
    \label{fig:obstacle_truck}
\end{figure}

To test the robustness of \algoName to outdated hints, we set up a goal-seeking experiment in one of the earlier environments and added a new obstacle---a large truck---blocking the path that \algoName took in the original trial. Since the satellite images are queried from a pre-recorded dataset, they do not reflect the addition of the truck, and hence continue to show a feasible path. We observe that the robot drives up to the truck and takes an alternate path to the goal, without colliding with it (see Figure~\ref{fig:obstacle_truck}). The lower-level latent goal model is robust to such obstacles and only proposes \emph{valid} subgoal candidates that do not lead to collision; since the learned heuristic only evaluates valid subgoals, \algoName is robust to small discrepancies in the hints.

\subsection{Incorrect Hints}
\label{sec:invalid_hints}

\begin{figure}
    \centering
    \includegraphics[width=\columnwidth]{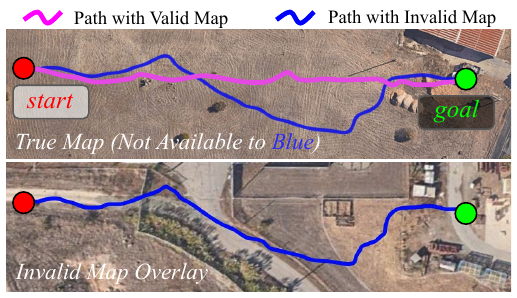}
    \caption{On navigation with invalid hints, like the map at a different location, \algoName deviates from its original path (magenta) and reaches the goal by following the learned heuristic (blue).}
    \label{fig:invalidmap}
    \vspace*{-0.1in}
\end{figure}

Next, we set up a goal-seeking experiment in one of the easy environments with modified GPS measurements, so that the satellite images available to \algoName are offset by a $\sim$5km constant. As a result, this \emph{hints} to the robot that there may be a road that it should follow, where in fact there isn't one (see Figure~\ref{fig:invalidmap}). We observe that the robot indeed deviates from its earlier path (with a valid map, the robot drives straight to the goal); upon overlaying this trajectory on the invalid map, we find that the learned heuristic indeed encourages the robot to follow the curvature of the road, but this path is still successful because it corresponds to open space. 

\subsection{A Disoriented \algoName}
\label{sec:ablations}

\begin{figure}
    \centering
    \includegraphics[width=\columnwidth]{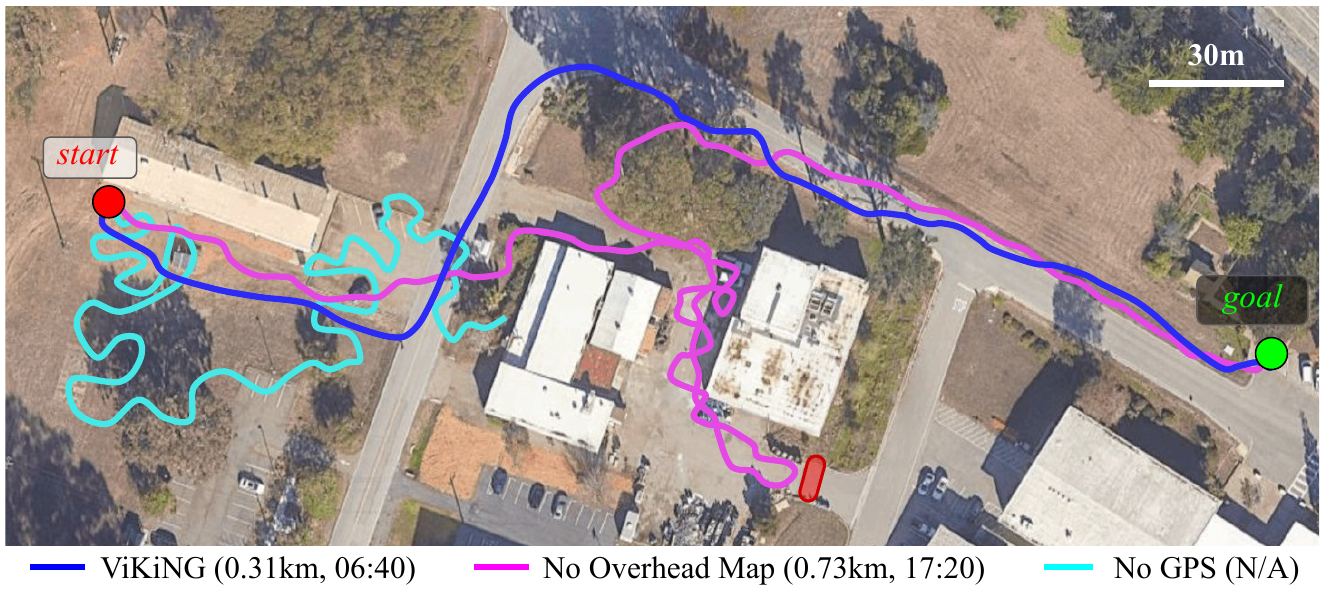}
    \caption{Ablations of \algoName by withholding geographic hints. \algoName without overhead images (magenta) acts greedily, driving close to buildings, gets caught into a cul-de-sac and eventually reaches the goal 2.6$\times$ slower that \algoName with access to satellite images (blue), which avoids the building cluster by following a smoother dirt path. Search without GPS (cyan) performs uninformed exploration and is unable to reach the goal in over 30 minutes.}
    \label{fig:ablations}
    \vspace*{-0.1in}
\end{figure}

Finally, we analyze the effects of \emph{disabling} the geographic hints and GPS localization on the goal-seeking performance of \algoName. Towards this, we run two variants of our algorithm:

    \noindent \textbf{No Overhead Image:} We provide the robot with GPS, but no satellite images. To accommodate this, we use a simple $\ell_2$ heuristic $h_\text{GPS}(x_w, x_g, x_t) = \|x_g - x_w\|$.
    
    \noindent \textbf{No GPS:} The robot does not have access to GPS or satellite images. To accommodate this, we remove the heuristic $h$ from \algoAstar, making it an uninformed search algorithm.

Figure~\ref{fig:ablations} summarizes the path taken by the robot, distance traversed, and time taken. When we disable the overhead hints and only use $h_\text{GPS}$, \algoAstar can still reach the destination, but takes significantly longer to do so, initially exploring a dead-end path that it then has to back out of. That said, this experiment also illustrates the ability of \algoAstar to handle less useful heuristics: while the path is significantly longer, the method is still able to eventually reach the destination, and in some sense the mistakes the method makes are to be expected of any system that has no prior map information. If we remove GPS as well, \algoAstar corresponds to a Dijkstra-like uninformed search (resembling RECON~\citep{shah2021rapid}). In this case, the robot searches its environment without any guidance and is unable to reach the goal in over 30 minutes.


\section{Discussion}
\label{sec:discussion}

We proposed a method for efficiently learning vision-based navigation in \emph{previously unseen} environments at a kilometer-scale. Our key insight is that effectively leveraging a small amount of geographic knowledge in a learning-based framework can provide strong regularities that enable robots to navigate to distant goals. We find that incorporating geographic hints as goal-directed heuristics for planning enables emergent preferences such as following roads or hiking trails. Additionally, \algoName only uses the hints for biasing the high-level search; the learned control policy at the lower-level relies solely on egocentric image observations, and is thus robust to imperfect hints. While we only use overhead images in our experiments, an existing avenue for future work is to explore how such a system could use other information sources, including paper maps or textual instructions, which can be incorporated into our contrastive objective.

\section*{Acknowledgments}
This research was partially supported by DARPA Assured Autonomy, ARL DCIST CRA W911NF-17-2-0181, DARPA RACER, and Toyota Research Institute. The authors would like to thank Blazej Osinski, Dieter Fox, Tambet Matiisen, Brian Ichter, and Katie Kang for useful discussions.

\IEEEpeerreviewmaketitle

\newpage
\bibliographystyle{IEEEtran} 
\bibliography{references}

\clearpage
\onecolumn
\appendix

\subsection{Implementation Details}
\label{app:implementation_details}

\begin{table*}[ht]
\centering
{\footnotesize
\begin{tabular}{llll}
\toprule
Layer & Input [Dimensions] & Output [Dimensions] & Layer Details \\ 
\midrule
\multicolumn{4}{c}{\emph{Encoder} $p_\phi(z \mid  o_t, o_w) = \mathcal N(\cdot ; \mu_p, \Sigma_p)$}  \\
\midrule
1 & $o_t$, $o_w$ [3, 160, 120] & $I_t^w$ [6, 160, 120] & Concatenate along channel dimension. \\
2 & $I_t^w$ [6, 160, 120] & $E_t^w$ [1024] & MobileNet Encoder \cite{howard2017mobilenets}\\
3 & $E_t^w$ [1024] & $\mu_p$ [64], $\sigma_p$ [64] & Fully-Connected Layer, $\mathrm{exp}$ activation of $\sigma_p$ \\ 
4 & $\sigma_p$ [64] & $\Sigma_p$ [64, 64] & $\mathrm{torch.diag}(\sigma_p)$  \\ 
\midrule
\multicolumn{4}{c}{\emph{Decoder} $q_\theta(a, d, x \mid  o_t, z_t^w) = \mathcal N(\cdot ; \mu_q, \Sigma_q)$}  \\
\midrule
1 & $o_t$ [3, 160, 120] & $E_t$ [1024]  & MobileNet Encoder \cite{howard2017mobilenets}\\
2 & $E_t$ [1024], $z_t^w$ [64] & $F=E_t\oplus z_t^w$ [1088] & Concatenate image and goal representation \\ 
3 & $F$ [1088] & $\mu_q$ [3], $\sigma_q$ [3] & Fully-Connected Layer, $\mathrm{exp}$ activation of $\sigma_q$ \\
4& $\sigma_q$ [5] & $\Sigma_q$ [5, 5] & $\mathrm{torch.diag}(\sigma_q)$  \\ 
5 & $\mu_q$ [5] & $\bar{a}_t^w [2], \bar{d}_t^w [1], \bar{x}_t^w [2]$ & Split into actions, distances and offsets\\
\bottomrule \\
\end{tabular}}
\caption{Architectural details of the latent goal model (Section~\ref{sec:representation})}\label{tab:recon_arch}
\end{table*}

\subsubsection{Latent Goal Model (Section~\ref{sec:representation})}
Inputs to the encoder $p_\phi$ are pairs of observations of the environment---current and goal---represented by a stack of two RGB images obtained from the onboard camera at a resolution of $160\times120$ pixels. $p_\phi$ is implemented by a MobileNet encoder~\cite{howard2017mobilenets} followed by a fully-connected layer projecting the $1024$-dimensional latents to a stochastic, context-conditioned representation $z_t^w$ of the goal that uses $64$-dimensions each to represent the mean and diagonal covariance of a Gaussian distribution. Inputs to the decoder $q_\theta$ are the context (current observation)---processed with another MobileNet---and $z_t^w$. We use the reparametrization trick~\cite{kingma2013auto} to sample from the latent and use the concatenated encodings to learn the optimal actions $a_t^w$, temporal distances $d_t^w$ and spatial offsets $x_t^w$. Details of our network architecture are provided in Table~\ref{tab:recon_arch}. During pretraining, we maximize $\gL_\text{VIB}$ (Eq.~\ref{eq:bottleneck_tractable}) with a batch size of 128 and perform gradient updates using the Adam optimizer with learning rate $\lambda=10^{-4}$ until convergence.

\subsubsection{Learned Heuristic (Section~\ref{sec:heuristic})}

Inputs to the encoder $p_\text{over}$ are (i) satellite image $c_S$ and (ii) the triplet of GPS locations $\{ x_w, x_S, x_G \}$. $p_\text{over}$ is implemented as a multi-input neural network with a MobileNet encoder~\cite{howard2017mobilenets} to featurize $c_S$, which is then concatenated with the location inputs. This is followed by a series of fully-connected layers [512, 128, 32, 1] down to a single cell to predict the binary classification scores. During pretraining, we minimize $\gL_\text{NCE}$ with a batch size of 256 and perform gradient updates using the Adam optimizer with learning rate $\lambda=10^{-4}$ until convergence.

\subsubsection{Miscellaneous Hyperparameters}
We provide the hyperparameters associated with our algorithms in Table~\ref{tab:hparams}.

\begin{table}[h!]
\centering
{\footnotesize
\begin{tabular}{clr}
\toprule
Hyperparameter & Value & Meaning \\ \midrule
$\Delta t$ & $0.5$ & Time step of the robot (s) \\
$\epsilon$ & $10$ & Threshold for $\mathrm{close}$ (Sec.~\ref{sec:search_primer})\\
$C$         & $20$ & Scaling constant for $v$ (Alg.~\ref{alg:exploration_algorithm} L\ref{algline:update_f})\\
$\lambda_\text{over}$ &  $200$  & Scaling constant for $h_\text{over}$ (Sec.~\ref{sec:heuristic}) \\
\bottomrule \\
\end{tabular}}
\caption{Hyperparameters used in our experiments.\label{tab:hparams}}
\end{table}

\subsection{Offline Trajectory Dataset}
\label{app:dataset}

For the offline dataset discussed in Section~\ref{sec:offline_dataset}, we use a combination of a 30 hours of autonomously collected data, and 12 hours of human teleoperated data. The complete dataset was collected by 3 independent sets of researchers over the course of 24 months in environments spanning multiple cities. We provide more information below.

\subsubsection{Autonomously Collected Data}
We use the published dataset by Shah et al.~\citep{shah2021rapid}, that contains over 5000 self-supervised trajectories collected over 9 distinct real-world environments. These trajectories capture the interaction of the robot in diverse environments, including phenomena like collisions with obstacles and walls, getting stuck in the mud or pits, or flipping due to bumpy terrain.

During data collection, a robot is equipped with a 2D LIDAR sensor to detect collisions ahead of time and generate autonomous pseudo-labels for collision events. To ensure that the control policy achieves sufficient coverage of the environment while also ensuring that the action sequences executed by the robot are realistic, we use a time-correlated random walk to gather data.

\subsubsection{Human Teleoperated Data}
The above dataset contains extremely diverse dataset that is great for learning general notions of traversability and collision avoidance. However, the random nature of the dataset means that it does not contain any semantically interesting behavior that may be desired of a robotic system, such as following a sidewalk or through a patch of trees. To enhance the quality of learned behaviors, we augment this dataset with about 12 hours of human teleoperated data in semantically rich environments such as hiking trails, city sidewalks, parking lots and suburban neighborhoods. These environments represent realistic scenarios where such a robotic system would be deployed.

\begin{figure}[h]
    \centering
    \includegraphics[width=0.8\columnwidth]{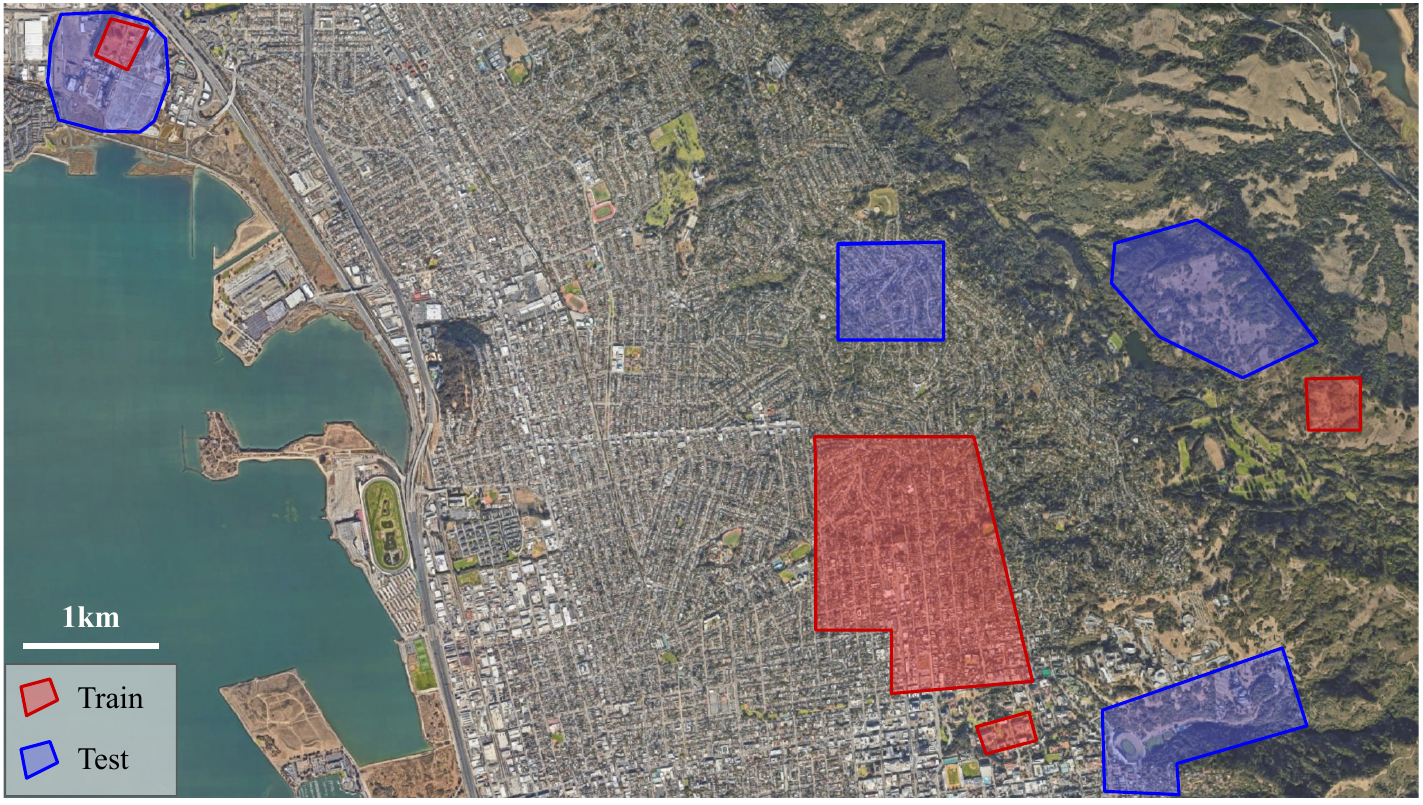}
\caption{Rough geographical locations of data collection by human teleoperation and testing (Section~\ref{sec:evaluation})}
    \label{fig:train_test}
\end{figure}

Table~\ref{tab:dataset_stats} summarizes key statistics of the trajectories, such as length and velocity. Table~\ref{tab:dataset_composition} summarizes the various environments in which the dataset was collected, and their relative composition. Figure~\ref{fig:train_test} visualizes the geographic locations of these data collection sites (location anonymized for the double-blind review process). We ensure no overlap between the training and test environments---success in these test environments  requires \emph{true} generalization to unseen environments.

\begin{table}
\centering
{\footnotesize
\begin{tabular}{lcc}
\toprule
 & Training Dataset & \algoName Deployment \\ \midrule
Avg. Length & 45m & $>$1km \\
Avg. Velocity (m/s) & 1.68 & 1.36 \\
\bottomrule
\end{tabular}}
\caption{Trajectory statistics for offline training dataset and real-world deployment.}
\label{tab:dataset_stats}
\end{table}

\begin{table}
\centering
{\footnotesize
\begin{tabular}{lc}
\toprule
Environment Type & Amount of Data (hrs) \\ \midrule
Paved Hiking Trails & 01:45 \\
City Sidewalks & 02:15 \\
Suburban Neighborhood Roads & 01:30 \\
Unpaved Grasslands & 01:00 \\
University/Office Campus & 02:30 \\
Miscellaneous & 03:00 \\ \midrule
\textbf{Total} & \textbf{12:00} \\
\bottomrule
\end{tabular}}
\caption{Approximate composition of various environment types in the teleoperated dataset.}
\label{tab:dataset_composition}
\end{table}

\subsection{Project Page}
We share experiment videos, including third-person perspectives of trajectories traversed by \algoName, on our project page: \href{https://sites.google.com/view/viking-release}{\bf\texttt{sites.google.com/view/viking-release}}.

\end{document}